%% file: main.tex
\pdfoutput=1

\documentclass[11pt]{article}

\usepackage[]{EACL2023}

\usepackage{times}
\usepackage{latexsym}

\usepackage[T1]{fontenc}

\usepackage[utf8]{inputenc}

\usepackage{microtype}

\usepackage{inconsolata}

\usepackage{booktabs}

\usepackage{adjustbox}
\usepackage{multirow}

\title{
Incorporating Question Answering-Based Signals into Abstractive Summarization via Salient Span Selection}

\author{Daniel Deutsch$^{\dagger}$ \\
  Google \\
  \texttt{dandeutsch@google.com} \\\And
  Dan Roth \\
  University of Pennsylvania \\
  \texttt{danroth@seas.upenn.edu} \\}

\newif\ifcomments
\commentstrue
\ifcomments
    \providecommand\dd[1]{\textcolor{blue}{[DD: #1]}}
    \providecommand\dr[1]{\textcolor{blue}{[DR: #1]}}
    \providecommand\todo[1]{\textcolor{red}{[TODO: #1]}}
\else
    \providecommand{\dd}[1]{}
    \providecommand{\dr}[1]{}
    \providecommand{\todo}[1]{}
\fi

\newcommand{\qaeval}{QA\-Eval}
\newcommand{\bertscore}{BERT\-Score}

\newcommand\blfootnote[1]{%
  \begingroup
  \renewcommand\thefootnote{}\footnote{#1}%
  \addtocounter{footnote}{-1}%
  \endgroup
}

\begin{document}
\maketitle
\input{00_abstract}
\blfootnote{$^{\dagger}$Work done while at the University of Pennsylvania}
\input{01_introduction}

\input{02_qa_supervision}

\input{03_model}

\input{04_data_augmentation}
\input{05_experimental_setup}
\input{06_results}
\input{07_related_work}
\input{08_conclusion}

\input{09_limitations}

\bibliography{bibliography}
\bibliographystyle{acl_natbib}

\appendix

\input{appendix/dataset_info}
\input{appendix/implementation_info}
\input{appendix/classifier_results}
\input{appendix/xsum}
\input{appendix/gold_span_annotation}
\input{appendix/augmented_comparison}
\input{appendix/human_eval_details}

\end{document}

%% file: 00_abstract.tex
\begin{abstract}
In this work, we propose a method for incorporating question-answering (QA) signals into a summarization model.
Our method identifies salient noun phrases (NPs) in the input document by automatically generating wh-questions that are answered by the NPs and automatically determining whether those questions are answered in the gold summaries.
This QA-based signal is incorporated into a two-stage summarization model which first marks salient NPs in the input document using a classification model, then conditionally generates a summary.
Our experiments demonstrate that the models trained using QA-based supervision generate higher-quality summaries than baseline methods of identifying salient spans on benchmark summarization datasets.
Further, we show that the content of the generated summaries can be controlled based on which NPs are marked in the input document.
Finally, we propose a method of augmenting the training data so the gold summaries are more consistent with the marked input spans used during training and show how this results in models which learn to better exclude unmarked document content.\footnote{
    \url{http://cogcomp.org/page/publication_view/997}
}

\end{abstract}

%% file: 01_introduction.tex
\section{Introduction}
Abstractive sequence-to-sequence summarization models have become very effective methods of easily generating summaries of input documents \citep{RMCW15,NZSGX16,LLGGMLSZ20}.

Previous work has demonstrated that conditioning the summary generation on salient document sentences results in higher-quality summaries and more controllable summarization models \citep{ChenBa18,DLHJN21}.
Salient sentences are typically identified during training by lexical overlap with the gold summaries \citep{NallapatiZhZh17} and predicted during inference.

Although marking different sentences as salient allows for some controllability over the content of the summary, desired summary content cannot be specified at the sub-sentence level.
Further, labeling sentences as salient via $n$-gram overlap does not directly take the predicate-argument structure of the text into account, which could result in a lower-quality supervision signal that misidentifies which particular instance of an $n$-gram is salient.

In this work, we propose to condition the summary generation on salient sub-sentence level spans which are identified by reasoning about the predicate-argument relations in the text.

\input{figures/intro-figure/intro_figure}

We mark noun phrases (NPs) in the input document as salient if the predicate-argument relation they participate in is present in the gold summary (\S\ref{sec:qa_supervision}).
This idea is implemented using automatic question generation (QG) and answering (QA).
For each NP, a wh-question that is answered by the NP is generated from the text.
Then, the NP is marked as salient if the generated wh-question is correctly answered in the gold summary according to a learned QA model, resulting in a more precise, sub-sentence level supervision signal (see Fig.~\ref{fig:intro_fig}).

The QA-based salience signal is incorporated into a two-stage summarization model (\S\ref{sec:model}).
First, a phrase salience classifier is trained to identify which NPs in the document are salient.
Then, the predicted salient spans are marked in the input document with special tokens and used to conditionally generate a summary of the document with a fine-tuned BART model \citep{LLGGMLSZ20}.

While we show that marking NPs as salient controls the summary content, the model often outputs extra, undesired information.
To that extent, we propose a data augmentation procedure that removes sentences unsupported by any salient span and generates new training examples based on what content should be able to be generated by subsets of the salient spans (\S\ref{sec:data_augmentation}).

Our experiments on three different summarization datasets show that the two-stage model trained with QA-based salient span supervision generates higher-quality summaries than lexical baseline methods of identifying salient spans on more extractive datasets according to several automatic evaluation metrics (\S\ref{sec:e2e_eval}).
Further, our data augmentation procedure results in summaries that are significantly shorter with only a small reduction in the percent of target content covered, demonstrating it successfully eliminates undesired summary content (\S\ref{sec:controllability_results}).



The contributions of our work include:
(1) a novel method of including QA-based signals into summarization generation;
(2) a two-stage model for incorporating phrase-level supervision into a summarization system;
and (3) a data-augmentation procedure which results in more controllable summarization models.

%% file: figures/intro-figure/intro_figure.tex
\begin{figure}
    \centering
    \includegraphics{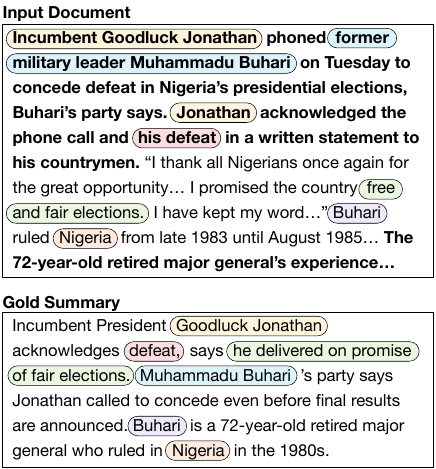}
    \caption{Salient spans identified by QA-based signals (shown in color) more precisely identify salient document content than those that identify salient sentences based on lexical overlap  (shown in bold).
    Our method classifies the salient spans, marks them in the input document, and then generates a summary.}
    \label{fig:intro_fig}
\end{figure}

%% file: 02_qa_supervision.tex
\section{Question-Based Salience}
\label{sec:qa_supervision}
We begin by describing how QA is used to identify salient spans of text in the input document and discuss the advantages of this approach.

We define a document NP as salient if its corresponding predicate-argument relation also appears in the gold summary.
To identify such NPs automatically, we employ question-generation and question-answering models as follows.

For each NP in the source document, we use the sentence it appears in to automatically generate a wh-question for which the NP is the answer.
This QA pair represents the predicate-argument relation that the NP participates in.
Then, we assume if a second text can be used to correctly answer that question, it contains the same predicate-argument relation.
Thus, we use a QA model to automatically answer the question against the gold summary and mark the NP as salient if the QA model predicts the question is answerable and the predicted answer is correct.
In practice, we assume a predicted answer is correct if it shares at least one token in common with the NP which was used to generate the question.

\input{figures/example-labeling/example_labeling}

An example of this procedure is illustrated in Fig.~\ref{fig:example_labeling} for two occurrences of the NP ``Sierra Leone.''
Questions for each phrase are automatically generated from the input document and answered against the gold summary.
Since the QA model correctly answered the first question but predicted the second question is not answerable, only the first occurrence of ``Sierra Leone'' is marked as salient.

We refer to the NPs identified by this procedure as  ``silver spans.''\footnote{
The term ``silver'' refers to the fact that the salient spans are not perfect because they were identified by sequence of learned models rather than humans (``gold'' spans; \S\ref{sec:e2e_eval}).}
Specific implementation details of the generation and answering models can be found in \S\ref{sec:setup}.

\subsection{Advantages of a QA-Based Approach}

Using QA to identify salient spans of text has several advantages.
First, because our QA approach operates at the phrase-level, it is able to be more precise about what specifically is salient in the document in contrast to sentence-level approaches.
For example, in the second sentence of Fig.~\ref{fig:intro_fig}, the QA-based salience signal identifies ``Jonathan'' and ``his defeat'' as salient but not ``written statement.''
A sentence-level approach would mark the entire sentence as salient and thus cannot make that distinction.

Second, because the QA-based approach reasons about the predicate-argument structure of the text, it is able to distinguish between which specific instances of the same NP are salient and which are not.
This is illustrated in Fig.~\ref{fig:example_labeling} in which the first occurrence of ``Sierra Leone'' is marked as salient but the second is not because the gold summary does say the health care worker was infected in Sierra Leone, but it does not say it is one of the hardest hit countries.
A salience signal that uses a bag-of-$n$-grams approach (e.g., ROUGE-based methods) cannot easily decide which instance ``Sierra Leone'' is salient.

%% file: figures/example-labeling/example_labeling.tex
\begin{figure}
    \centering
    \includegraphics{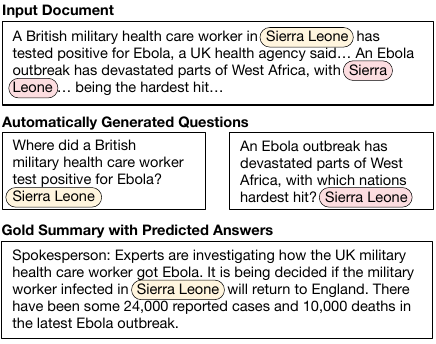}
    \caption{
    We define a document noun phrase as salient if the wh-question it answers is also answered in the gold summary.
    Here, the first (yellow) instance of ``Sierra Leone'' is salient and the second (red) is not because the gold summary answers the automatically generated question for the first instance but not the second.
    }
    \label{fig:example_labeling}
\end{figure}

%% file: 03_model.tex
\section{A Two-Stage, Span-Based Model}
\label{sec:model}

Next, we propose a two-stage, span-based model that can incorporate the QA-based salience signals into the learning procedure.
The first of the two stages, the span selection component, classifies salient spans within the text.
The second stage, the generation component, generates the summary given the document and the salient spans.
The details of each component are detailed next.

\subsection{Salient Span Classifier}
Given an input document $d = [x_1, \dots, x_n]$ and a set of spans $S$, in which each span $s_{i,j}$ represents a sequence of tokens $x_i, \dots, x_j$ in $d$, the span classifier outputs a score for each span based on how salient it is in the document.
Our definition of salience is discussed in \S\ref{sec:qa_supervision}.

Concretely, the input tokens are first encoded using BART.
Then, the representation of a span is created by concatenating the BART encodings of the first and last tokens in the span.
Finally, a linear classifier is trained using this encoding to predict the salience of each span.

A set of silver spans $S^* \subseteq S$ is used to train the model using a binary cross-entropy loss.
When using the QA-based approach, $S$ is the set of NPs in the document and $S^*$ is the subset that our QG-QA algorithm identified as salient.
We reweight the loss term of each span such that positive and negative spans contribute equally.
During inference, a score is predicted for each span in $S$ and the top-$k$ sorted by highest score are passed to the generation component.
We choose the $k$ spans independently, although they could also be selected jointly.

\input{figures/augmentation-example/augmentation_example}

\subsection{Generation Component}
Given an input document and set of salient spans, the generation component produces a summary of the document.
The salient spans are represented by inserting special tokens directly into the document's sequence of tokens before and after the spans.
For example, if span $s_{4,5}$ was marked as salient, the document's tokens would be represented as
\begin{verbatim}
  ... x3 [SS] x4 x5 [SE] x6 ...
\end{verbatim}
where \texttt{[SS]} and \texttt{[SE]} mark the start and end of the span.

Since the salient spans are represented in the document tokens, we are able to directly train a sequence-to-sequence model to generate the gold summary from the modified document representation without any changes to the model's architecture.

During training, we use silver spans and the ground-truth summary to fine-tune BART using a standard cross-entropy loss function.
The ground-truth summary does not have any marking of salient spans.
For inference, the predicted salient spans from the span classifier are used instead of the silver spans.

%% file: figures/augmentation-example/augmentation_example.tex
\begin{figure*}
    \centering
    \includegraphics{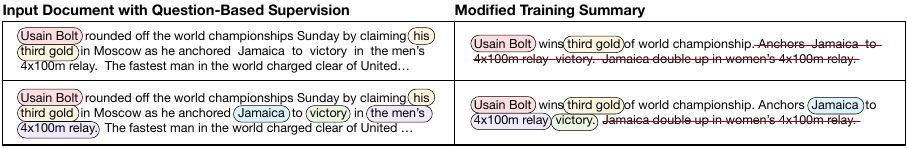}
    \caption{
    An example of our data augmentation procedure.
    The colors represent the mapping between document and summary spans.
    The document spans are given to the generation model during training.
    In this example, no span maps to the third summary sentence, so it is removed entirely.
    Then, new training instances are generated using the first summary sentence and first two summary sentences with their corresponding salient document spans.
    }
    \label{fig:augmentation_example}
\end{figure*}

%% file: 04_data_augmentation.tex
\section{Improving Controllability via Data Augmentation}
\label{sec:data_augmentation}

Although there is nothing to directly force the generation model to learn to include content based on the supervision provided by the salient spans, if the supervision is of high enough quality, we expect the model will learn to do so.
Indeed, we later show in \S\ref{sec:controllability_results} that this is true, thus the content of the summary can be controlled by which spans are marked as salient.
However, it is also desirable for a controllable summarization model to also not include content which is not marked as salient.
The generation models may learn to include extra information for at least two reasons.

First, the gold summaries may include content which cannot be generated based on only the silver salient spans that were used to train the generation model, so it may learn to output extra, unmarked information.
This could happen if the QG/QA models are imperfect (resulting in a noisy supervision signal) or if the gold summary contains information that cannot be mapped to the document.
Second, if the model is trained to generate summaries of a certain length and the length of the summary necessary to include all of the information marked by the spans is smaller than those used for training --- for example, because the number of marked spans is small --- the model could generate additional information simply to increase the summary length.

An artifact of our silver span annotation procedure enables us to address these controllability issues.
If a document span is marked as salient, that means it has a corresponding phrase in the gold summary which expresses the same content.
Therefore, the QG-QA procedure creates a mapping between which parts of the gold summary should be able to be generated by marking different parts of the input document.

We propose to leverage this mapping to augment the training data in two ways.
First, we remove any gold summary sentence which has no phrase mapped to the document.
These sentences would encourage the model to generate additional content based on unmarked spans.

Second, we generate new pairs of salient spans and gold summaries for training by selecting the first $k$ remaining gold summary sentences and the subset of salient document spans which map to them.
For instance, if $k=2$, only the salient spans which are mapped to the first two summary sentences are marked in the input document, and the model is trained to generate only those sentences.
We generate new examples for each original training instance using all possible values of $k$.
By training on these new pairs, the model should learn to better control the length of the output summary based on the number of marked salient spans.
An example of these augmentations is included in  Fig.~\ref{fig:augmentation_example}.

Although this procedure is described within the context of the QA-based supervision, it can be implemented with any such mapping between the document and gold summaries. 

%% file: 05_experimental_setup.tex
\section{Experimental Setup}
\label{sec:setup}

\paragraph{Datasets}
Our experiments use three popular English single-document summarization datasets: CNN/DailyMail \citep{NZSGX16}, XSum \citep{NarayanCoLa18a}, and NYTimes \citep{Sandhaus08}.
Specific details on the sizes of the datasets can be found in Appendix~\ref{app:dataset_stats}.

\paragraph{Baselines \& Other Work}
We compare the salient spans selected by our QA-based method against three baseline span selection methods.
The first marks salient sentences by greedily selecting $k$ sentences that maximize the ROUGE-2 score calculated against the gold summary, a popular method that is frequently used to train extractive summarization models \citep{NallapatiZhZh17} as well as other two-step abstractive systems \citep{ChenBa18,DLHJN21}.
The other two mark entities and  NPs as salient if they appear in the gold summaries as determined by lexical matching.
We only mark the first occurrence of the phrases as salient since we found that worked better than marking all occurrences.

Additionally, we compare our results to BART \citep[the original implementation and our own;][]{LLGGMLSZ20} since our models are built on top of it.
We also compare to GSum \citep{DLHJN21}, which uses salient sentence guidance that is similar to our baseline salient sentence method.
GSum encodes the additional guidance signal separately from the input document and uses the document and guidance encodings to generate the summary.

\paragraph{Summarization Evaluation Metrics}
The models are automatically evaluated using three metrics which calculate a similarity score between the generated and gold summaries.
ROUGE \citep{Lin04} compares the two summaries based on their lexical overlap.
\bertscore{} \citep{ZKWWA20} calculates a similarity score between the summaries based on their tokens' BERT embeddings \citep{DCLT19}.
\qaeval{} \citep{DeutschBeRo21} is a QA-based evaluation metric which generates questions from the gold summaries and answers them against the generated summaries.
Its similarity score is equal to the average token F$_1$ score calculated between the predicted and expected answers.

We additionally perform a human evaluation of summary quality on Mechanical Turk.
We ask 3 Turkers to rate the quality of 50 summaries per model from the CNN/DailyMail dataset on a scale from 1 to 5 based on the importance of the information, faithfulness, fluency, and coherence.
Details on the manual evaluation can be found in Appendix~\ref{app:human_eval}.

\paragraph{Controllability Evaluation Metrics}
The controllability of our model is evaluated using the \emph{question recall}.
Given $k$ marked spans, we define the question recall to be equal to the percent of the corresponding $k$ wh-questions that are answered by the summary according to the QA model.
This approximates the recall on the desired predicate-argument structures in the summary.
We additionally report the ratio between $k$ and the length of the generated summary in tokens to measure the precision of the generated information.
A larger value means the summary is more concise.

\paragraph{Implementation Details}
The QG/QA models are the same as used by \qaeval{}.
The generation model is initialized with BART-Large and fine-tuned on data collected by \citet{DemszkyGuLi18}.
The answering model is initialized with ELECTRA-Large \citep{CLLM20} and fine-tuned on SQuAD 2.0 \citep{RajpurkarJiLi18}.

The span classification and generation models are both initialized with BART-Large and fine-tuned on the respective datasets.
They were trained for three and five epochs, respectively, and the model with the best precision@1 and ROUGE-2 F$_1$, respectively, on the validation set were selected as the final models.
See Appendix~\ref{app:implementation} for more specific implementation details.

%% file: 06_results.tex
\section{Results}
\label{sec:results}

\subsection{Summarization Evaluation}
\label{sec:e2e_eval}

\paragraph{Automatic Evaluation}
Table~\ref{tab:automatic_results} contains the models' performances as evaluated by automatic metrics, both using the spans predicted by the classifier (``end-to-end'') and the silver spans (i.e., assuming a ``perfect'' classifier).

\input{figures/automatic_results}

Interestingly, we find a somewhat surprising result.
On CNN/DailyMail and NYTimes, the end-to-end QA-based model performs the best among the different span labeling methods and the baseline BART.
On NYTimes, it is also better than GSum.
However, if the silver span labels are used, the lexical NP-based model out-performs the rest by a somewhat large margin.
It is surprising that a seemingly better generation model would result in worse end-to-end performance.

To better understand this result, we manually labeled all of the NPs in 50 CNN/DailyMail documents as salient or not salient based on whether the corresponding predicate-argument relation was present in the reference summary (see Appendix~\ref{appendix:gold_span} for details).
These spans, which we call the gold spans, can be used to evaluate the precision and recall of the silver spans as well as the output from the salient span classifiers.

\input{figures/supervision_eval}

Table~\ref{tab:silver_pr} shows that the QA-based labels are more precise but have lower recall than the lexical NP labels.
Because the lexical NP method aggressively marks the first occurrence of any NP in the document which is present in the reference as salient, it is unsurprising that its recall would be high.
Since it cannot distinguish between instances of the same NP due to its bag-of-words representation, its precision is low.
In contrast, the QA-based approach can reason about which occurrence of an NP is salient (resulting in higher precision), but the recall is lower likely due to noise in the QG/QA models.
This same pattern appears in Table~\ref{tab:silver_pr} with the outputs from the salient span classifiers, although the precisions and recalls are notably lower than the silver span labels'.

We believe that the discrepancy between the end-to-end and silver span-based models' performances can be explained by these results.
The lexical NP generation model was trained with a high recall silver supervision at 66.3, allowing the generation model to achieve good performance when the silver spans are provided.
Yet during inference the model is provided with spans that only have around 54.4 recall, 12 points lower.
We suspect the generation model learned to rely heavily on the marked salient spans --- and empirically we observed that it copied very heavily from them --- thus when the quality of the span signal was reduced, the resulting summaries similarly got worse.
In contrast, the difference between the QA-based model's recall during training and inference is only estimated to be around 2.4, so this issue is less severe, resulting in better end-to-end summaries.

To test this hypothesis, we artificially ablated the lexical NP-based generation model's silver span supervision's recall by removing $k$\% of the salient spans uniformly at random --- thus making the training spans look more similar to the spans during inference --- and retrained the model.
We would expect the silver span-based model's performance to decrease while the end-to-end model's increases.
Indeed, in Table~\ref{tab:ablation_results} we find that this does happen.
These results suggest that the relationship between the classifier's performance and generation model's supervision is important for good end-to-end results and could be explored in future work.

\input{figures/ablation_results}

Although the end-to-end lexical NP results begin to approach the QA-based model's performance, they do not quite reach it.
Further, the QA-based silver spans maintain an F$_1$ advantage over the lexical NP method (Table~\ref{tab:silver_pr}).
While the QA-based approach can be improved with better question generation and answering models, the lexical NP labeling method is inherently limited.
Therefore, the QA-based method does appear to be the best method of incorporating additional supervision into the summarization models based on the automatic metrics.

\paragraph{Human Evaluation}
Table~\ref{tab:human_eval_quality} contains the results of evaluating BART and the sentence- and QA-based models on CNN/DailyMail (the best performing) using human summary quality annotations from Mechanical Turk.
On average, our span-based methods have higher quality summaries than the baseline method of BART.
After collecting annotations for 50 summaries on CNN/DailyMail, we were unable to obtain statistical significance between the two span-based models, however, doing so may be prohibitively difficult \citep{WeiJi21}.

\input{figures/human_eval_quality}

\subsection{Controllability Evaluation}
\label{sec:controllability_results}

\paragraph{Automatic Evaluation}
The controllability of the QA-based generation model is evaluated in Fig.~\ref{fig:controllability_pr} using the original training data as well as the augmented data described in \S\ref{sec:data_augmentation}.
We plot the question recall and the ratio between $k$ and the length of the generated summaries for the top $k$ most salient spans output by the QA-based salient span classifier for various values of $k$ on CNN/DailyMail.
The data augmentation procedure is split into only removing sentences that do not answer a question (``+Rm Sents'') plus also generating new training examples (``+New Examples'').
We also include the results for BART (for which the summary is constant for all $k$) for relative comparisons.

\input{figures/controllability-plot/controllability_plot}

\input{figures/controllability-example/controllability_example}

Although BART's question recall is initially higher than the QA models' recalls, as $k$ increases it falls lower.
We suspect this is because BART has learned to include the same content that the span classifier also identifies as salient when $k$ is small and the length of its summaries allows it to cover more content.
However, when $k$ increases, the span classifier potentially predicts different spans as salient than what BART learned, resulting in divergent content and a lower recall for BART.
The higher recall of the QA models demonstrates that their summary content is indeed being controlled via the input spans.
Further, the QA models have far better $k$-to-length ratios, meaning their summaries are shorter than BART's even when their recalls are higher, suggesting they generate far less content which is unrelated to the marked spans.

Among the QA-based models, we do observe a small drop in recall when the model is trained with data augmentation.
However, the data-augmented summaries express that information far more concisely (because the ratio between $k$ and the summary length is higher).
For example, when 10 input spans are marked, there is a relative 0.9\% and 3.2\% drop in recall for removing sentences and the full data augmentation procedure, respectively, but the summary lengths are 14\% and 22\% shorter.
Therefore, the data augmentation procedures do result in models which have learned to not generate extra content.

\paragraph{Controllability Example}
Example summaries from the QA models and sentence-based model with different marked input spans are shown in Fig.~\ref{fig:controllability_example}.
Because the sentence-based model is limited to marking full sentences, the content which is taken from the marked sentence cannot be further controlled.
In contrast, the figure shows how the QA models' summaries can be altered by marking different NPs within the sentence, thus demonstrating the benefits of phrase-level controllability.

The example in Fig.~\ref{fig:controllability_example} also shows how the data augmentation procedure improves controllability.
The phrases which the standard model includes but the augmented model does not are marked in bold.
The augmented model does a better job at excluding content which was not marked in the input document.

%% file: figures/automatic_results.tex
\begin{table*}
    \centering
    \begin{adjustbox}{width=0.9\textwidth}
\begin{tabular}{lccccccccccc}
\toprule
\multirow{2}{*}[-0.2em]{\bf Method} & \multicolumn{5}{c}{\bf CNN/DailyMail} && \multicolumn{5}{c}{\bf NYTimes} \\
\cmidrule{2-6} \cmidrule{8-12} 
& \bf R1 & \bf R2 & \bf RL & \bf BSc & \bf QAE & & \bf R1 & \bf R2 & \bf RL & \bf BSc & \bf QAE \\
\midrule
\multicolumn{12}{l}{\emph{Baselines \& Other Work}} \\
BART & 44.2 & 21.3 & 40.9 & - & - & & - & - & - & - & -\\
BART (ours) & 44.1 & 21.0 & 40.9 & 88.3 & 23.5 & & 54.0 & 35.2 & 50.7 & 89.5 & 27.3\\
GSum & \hphantom{$^\dagger$}46.0$^\dagger$ & \hphantom{$^\dagger$}22.3$^\dagger$ & \hphantom{$^\dagger$}42.6$^\dagger$ & \hphantom{$^\dagger$}88.6$^\dagger$ & 22.9 & & 54.3 & 35.4 & 47.6 & - & -\\
\midrule
\multicolumn{12}{l}{\emph{Silver Spans}} \\
Sentences & 51.7 & 29.9 & 48.8 & 89.4 & 28.6 & & 62.7 & 46.0 & 59.8 & 91.2 & 33.5\\
Entities & 51.5 & 27.6 & 48.0 & 89.6 & 30.0 & & 60.9 & 42.8 & 57.6 & 90.8 & 32.0\\
Lexical NPs & \bf 59.6 & \bf 34.6 & \bf 55.8 & \bf 90.6 & \bf 36.2 & & \bf 68.2 & \bf 50.7 & \bf 64.8 & \bf 92.0 & \bf 36.6\\
QAs & 55.3 & 31.4 & 51.9 & 90.0 & 33.7 & & 65.7 & 48.7 & 62.6 & 91.6 & 35.8\\
\midrule
\multicolumn{12}{l}{\emph{End-to-End}} \\
Sentences & 45.0 & \bf 21.8 & 41.8 & 88.2 & 23.2 & & 54.6 & 35.9 & 51.4 & 89.6 & 27.6\\
Entities & 43.5 & 20.3 & 40.4 & 88.3 & 23.2 & & 53.5 & 34.6 & 50.3 & 89.4 & 27.0\\
Lexical NPs & 44.8 & 21.0 & 41.6 & 88.4 & 23.2 & & 54.6 & 35.4 & 51.3 & 89.6 & 27.1\\
QAs & \bf 45.5 & \bf 21.9 & \hphantom{$^\dagger$}\bf 42.4$^\dagger$ & \bf 88.5 & \hphantom{$^\dagger$}\bf 24.4$^\dagger$ & & \hphantom{$^\dagger$}\bf 55.2$^\dagger$ & \hphantom{$^\dagger$}\bf 36.3$^\dagger$ & \hphantom{$^\dagger$}\bf 51.9$^\dagger$ & \hphantom{$^\dagger$}\bf 89.7$^\dagger$ & \hphantom{$^\dagger$}\bf 28.0$^\dagger$\\
\bottomrule
\end{tabular}

    \end{adjustbox}

    \caption{The automatic metric results for the baselines and other work (top), models that use silver spans (middle), and end-to-end models (bottom) evaluated with ROUGE (R1, R2, RL), \bertscore{} (BSc), and \qaeval{} (QAE).
    Values in bold are statistically the best in each section and $\dagger$ marks the best values overall (excluding silver labels) using a permutation test with $\alpha = 0.05$.
    }
    \label{tab:automatic_results}
\end{table*}

%% file: figures/supervision_eval.tex
\begin{table}
    \centering
    \small
    \begin{tabular}{lccc}
        \toprule
        \bf Method & \bf Precision & \bf Recall & \bf F$_1$ \\
        \midrule
        \multicolumn{4}{l}{\emph{Silver Labels}} \\
        Lexical NPs & 32.7 & \bf 66.3 & 41.8 \\
        QAs & \bf 43.8 & 51.5 & \bf 45.3 \\
        \midrule
        \multicolumn{4}{l}{\emph{Predicted Spans}} \\
        Lexical NPs@25 & 23.8 & \bf 54.4 & \bf 32.0 \\
        QAs@20 & \bf 27.3 & 49.1 & \bf 33.8 \\
        \bottomrule
    \end{tabular}
    \caption{The average summary-level precision, recall, and F$_1$ scores of the silver labeling methods (top) and the output from the span classifiers (bottom) evaluated against the human-annotated gold labeling.
    Results in bold are statistically higher (or tied) under a single-tail pairwise permutation test with $\alpha = 0.05$.
    The @$k$ values were selected based on validation set performance.
    }
    \label{tab:silver_pr}
\end{table}

%% file: figures/ablation_results.tex
\begin{table}
    \centering
    \begin{adjustbox}{width=0.6\columnwidth}
    \begin{tabular}{lccc}
        \toprule
\multirow{2}{*}[-0.2em]{\bf Method} & \multicolumn{3}{c}{\bf CNN/DailyMail} \\
\cmidrule{2-4}
& \bf R1 & \bf R2 & \bf RL \\
        \midrule
        \multicolumn{4}{l}{\emph{Silver Spans}} \\
        Lexical NPs & \bf 59.6 & \bf 34.6 & \bf 55.8 \\
        \quad +10\% Noise & 57.8 & 32.8 & 54.0 \\
        \quad +20\% Noise & 56.3 & 31.5 & 52.6 \\
        \quad +30\% Noise & 55.0 & 30.4 & 51.4 \\
        \quad +35\% Noise & 54.1 & 29.6 & 50.6 \\
        QAs & 55.3 & 31.4 & 51.9 \\
        \midrule
        \multicolumn{4}{l}{\emph{End-to-End}} \\
        Lexical NPs & 44.8 & 21.0 & 41.6 \\
        \quad +10\% Noise  & 45.0 & 21.3 & 41.8 \\
        \quad +20\% Noise  & 45.2 & 21.6 & 42.0 \\
        \quad +30\% Noise  & 45.3 & 21.7 & 42.1 \\
        \quad +35\% Noise  & 45.1 & 21.6 & 41.9  \\
        QAs & \bf 45.5 & \bf 21.9 & \bf 42.4 \\
        \bottomrule
    \end{tabular}

    \end{adjustbox}

    \caption{The ablated lexical NP supervision shows as the noise increases, the silver span performance decreases but end-to-end performance improves.
    }
    \label{tab:ablation_results}
\end{table}

%% file: figures/human_eval_quality.tex
\begin{table}
    \centering
    \small
    \begin{tabular}{lccc}
        \toprule
         & \bf BART & \bf Sentences & \bf QA \\
        \midrule
        Quality Score & 3.76 & \bf 3.86 & \bf 4.00 \\
        \bottomrule
    \end{tabular}
    \caption{Summary quality scores according to humans.
    Results in bold are statistically tied for the best score.}
    \label{tab:human_eval_quality}
\end{table}

%% file: figures/controllability-plot/controllability_plot.tex
\begin{figure}
    \centering
    \includegraphics[width=\columnwidth]{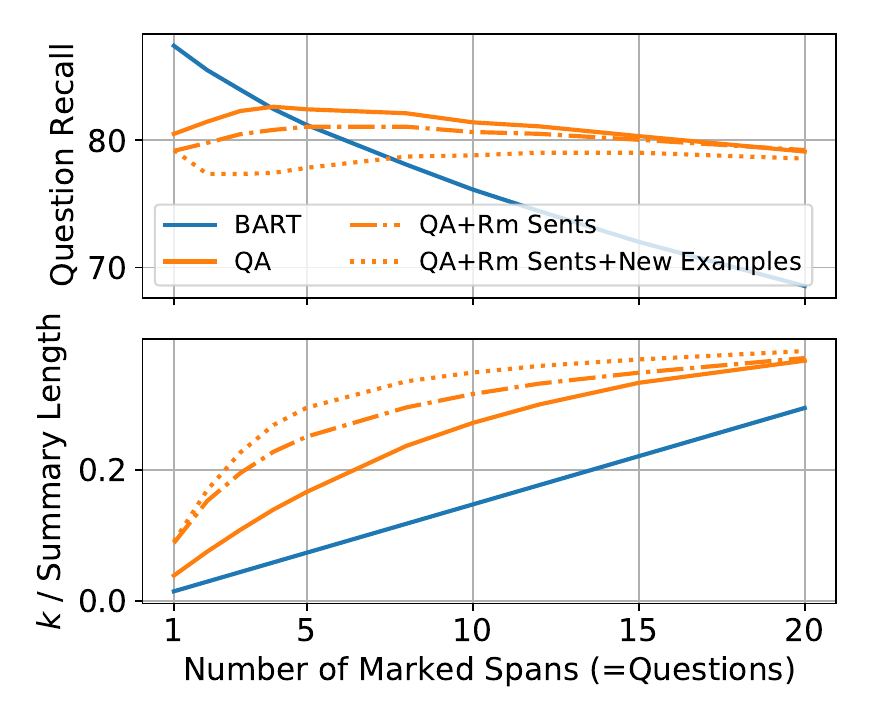}
    \caption{The percent of questions which correspond to the marked spans answered by the generated summaries (top) and the summary lengths in tokens (bottom).
    The QA methods have higher question recall than BART and are far more concise, demonstrating that marking input spans controls the summary content.
    }
    \label{fig:controllability_pr}
\end{figure}

%% file: figures/controllability-example/controllability_example.tex
\begin{figure*}
    \centering
    \includegraphics[width=\textwidth]{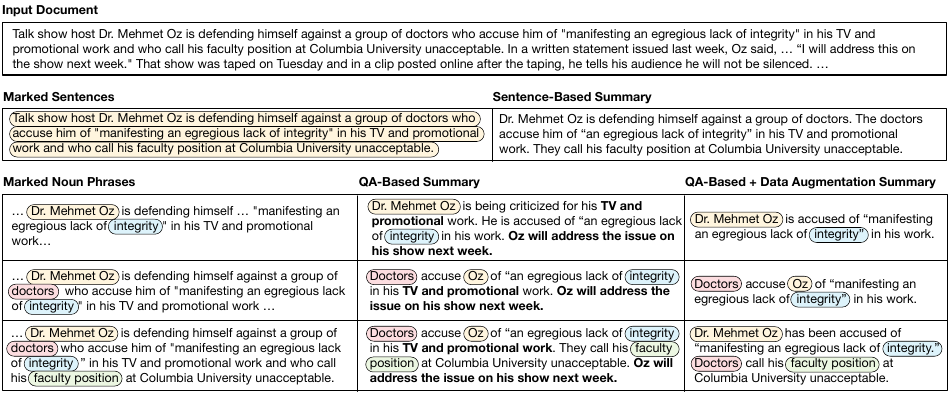}
    \caption{Example summaries generated by the sentence-based model (middle), QA-based model (bottom center) and QA-based model trained on the augmented data (bottom right).
    The QA-based models allow for much more control over the summary content than the sentence model by marking different combinations of phrases.
    The augmented-data summaries better eliminate unmarked content from the input than the standard model (extra information generated by the standard model shown in bold).
    }
    \label{fig:controllability_example}
\end{figure*}

%% file: 07_related_work.tex
\section{Related Work}
\label{sec:related_work}

\paragraph{QA-Based Signals}
QA-based signals have been used for evaluating summaries \citep{EyalBaEl19,DurmusHeDi20,WangChLe20,DeutschBeRo21}, including \citet{SDLPSWG21}, who explore a similar notion of document salience.
They have also been used to align content across documents \citep{WRKED21} as well as train summarization models \citep{ArumaeLi18,ArumaeLi19,SLPS19}.
The models which incorporate QA-based signals typically do so using reinforcement learning.
In contrast, our approach is simpler.
We incorporate the QA-based signal by marking spans in the document, and our models are trained using easier-to-optimize cross-entropy objective functions.

\paragraph{Incorporating Additional Supervision}
Recent work by \citet{DLHJN21} proposes a framework for incorporating additional guidance into summarization models, called GSum.
They separately encode the input document and the supervision signal, whereas we directly mark spans in the text.
This allows for our generation component to have a simpler architecture than theirs.
While they are able to encode any natural language string, our model provides more direct supervision by identifying which specific tokens are salient.

Other work has included predicate-argument structure into summarization to generate more faithful summaries \citep{CWLL18,JinWaWa20,ZHXZZHJ21}.
They represent the predicate-arguments either using dependency trees or OpenIE tuples, whereas we represent them via QA pairs.
These works include that information to try and generate faithful summaries, whereas our goal is to identify salient document content.

\paragraph{Controllable Summarization}
Work on controllable summarization has focused on aspects such as the length of the summary \citep{FanGrAu18} and the content in an interactive setting \citep{SRAABD17} or via prompting \citep{HKMRX20}.
Incorporating our QA-based signal via prompting may be difficult given the number of questions which would need to be concatenated onto the input.

Other approaches control content via planning as in entity templates \citep{NZMSNM21}, marking records in a data-to-text approach \citep{PuduppullyDoLa19}, or using aspect controllers \citep{AmplayoAnLa21}.
The marked salient spans in our work could be viewed as a content plan as well.

\paragraph{Data Augmentation}
Previous work has proposed methods for removing sentences or full summaries from the training data in order to discourage the summarization model from learning to generate unfaithful information \citep{MatsumaruTaOk20,NNWSZZMX21,NZMSNM21}.
In addition to removing sentences, we generate new training instances in order to learn to exclude content which is not marked as salient in the input, resulting in more controllable models.

%% file: 08_conclusion.tex
\section{Conclusion}
In this work, we proposed a method for incorporating QA-based signals into a summarization model by automatically marking document NPs as salient based on whether a NP's corresponding wh-question is answered correctly in the summary.
We showed that incorporating this signal into our two-stage summarization model results in higher quality summaries than baseline methods of identifying salient spans.
Finally, we demonstrated that our data augmentation algorithm, which attempts to ensure the span supervision is consistent with the gold summaries, improves controllability by eliminating unmarked content from the output summaries.

\section*{Acknowledgments}
The authors would like to thank the anonymous EACL reviewers for their insightful feedback on our work.

This work was supported by a Focused Award from Google and Contracts FA8750-19-2-1004 and FA8750-19-2-0201 with the US Defense Advanced Research Projects Agency (DARPA). Approved for Public Release, Distribution Unlimited. The views expressed are those of the authors and do not reflect the official policy or position of the Department of Defense or the U.S. Government.

%% file: 09_limitations.tex
\section*{Limitations}
The summarization model proposed in this work relies on the existence of question generation and question answering models, which are used to pre-process the data for training.
It is likely that such models only exist or are of high-enough quality for high resource languages, such as English, which limits the languages that our model can be trained on in practice.
However, we do not see a reason why our model could not be applied to another language as long as those additional resources are available.

Further, because the question generation model produces simple wh-questions and the question answering model is only able to reason about the predicate-argument structure of the text (due to it being trained on SQuAD 2.0), our procedure for identifying salient document phrases requires that the same information across the document and reference summary must be expressed in relatively similar ways (e.g., up to rephrasing and synonyms).
If the reference summary does contain the answer to the generated question but identifying that answer requires a level of reasoning beyond reasoning about the predicate-argument structure (e.g., does it require multi-hop reasoning?), the specific models proposed in this work may fail to identify those salient phrases.
This limits the types of datasets for which we expect our proposal to do well on (discussed more in Appendix~\ref{appendix:xsum}).
However, if matching document and reference summary information requires a level of reasoning that is supported by the generation and answering models, then we suspect our proposal will work, in theory, but we have not experimented with this in practice.

Our methods out-perform the baseline systems the most when evaluated by QA\-Eval, and we leverage the same QG/QA technology as this metric.
While this commonality may bias our system toward summaries that are favored by QA\-Eval, we argue this is not necessarily a bad thing.
Previous work has incorporated ROUGE-based signals into their models, either indirectly by selecting extractive labels based on ROUGE or directly by optimizing ROUGE via reinforcement learning.
Our approach is analogous to these modeling approaches, and QA\-Eval has been demonstrated to be a better evaluation metric than ROUGE, so it is likely a better metric to optimize for.

%% file: appendix/dataset_info.tex
\section{Dataset Statistics}
\label{app:dataset_stats}
The sizes of the CNN/DailyMail, XSum, and NYTimes datasets are included in Table~\ref{tab:dataset_stats}.
The Table also includes the number of spans per span type that were selected from the classification component and passed to the generation component during inference.
The values were selected based on a parameter sweep on the validation set.
The number of spans with the highest ROUGE-2 F$_1$ score was selected.

\input{figures/dataset_stats}

%% file: figures/dataset_stats.tex
\begin{table}
    \centering
    \begin{adjustbox}{width=\columnwidth}
    \begin{tabular}{cccccc}
        \toprule
        \bf Dataset & \bf \#Train & \bf \#Valid & \bf \#Test & \bf Span Type & \bf \#Spans \\
        \midrule
        \multirow{4}{*}{CNN/DM} & \multirow{4}{*}{287,113} & \multirow{4}{*}{13,368} & \multirow{4}{*}{11,490} & Sentences & 3 \\
        & & & & Entities & 10 \\
        & & & & Lex. NPs & 25 \\
        & & & & QA & 20 \\
        \midrule
        \multirow{4}{*}{XSum} & \multirow{4}{*}{204,045} & \multirow{4}{*}{11,332} & \multirow{4}{*}{11,334} & Sentences & 1 \\
        & & & & Entities & 1 \\
        & & & & NPs & 5 \\
        & & & & QA & 1 \\
        \midrule
        \multirow{4}{*}{NYTimes} & \multirow{4}{*}{44,382} & \multirow{4}{*}{5,523} & \multirow{4}{*}{6,495} & Sentences & 4 \\
        & & & & Entities & 15 \\
        & & & & Lex. NPs & 45 \\
        & & & & QA & 27 \\
        \bottomrule
    \end{tabular}
    \end{adjustbox}
    \caption{The number of instances in the training, validation, and test splits of the three datasets used in our experiments as well as the number of spans selected by the classification component that were passed as input to the generation component.}
    \label{tab:dataset_stats}
\end{table}

%% file: appendix/implementation_info.tex
\section{Implementation Details}
\label{app:implementation}
All of the models were trained with the same hyperparameters for across datasets and span types which were based on those used by BART \citep{LLGGMLSZ20}.

The classification component was a BART-Large model that was fine-tuned with a binary cross-entropy classification loss.
We selected the model based on which had the best precision@1 on the validation dataset.
The generation models were also fine-tuned BART-Large models, but they instead use a cross-entropy loss function.

Both the components were trained using Adam \citep{LoshchilovHu19} with weight decay and learning rate 3e-5.
The classification component was trained for 3 epochs, and the final model was selected based on the precision@1 on the validation set.
The generation component was trained for 5 epochs, and the final model was selected based on the ROUGE-2 F$_1$ score on the validation set.

%% file: appendix/classifier_results.tex
\section{Salient Span Classifier Evaluation}
\label{sec:classifier_eval}

Fig.~\ref{fig:classifier_pr} contains the precision@$k$ and recall@$k$ of the span based classifiers calculated against the corresponding silver spans.
These plots should be interpreted as how well the span classifiers were able to learn from their respective supervision, not necessarily the true quality of the output span labels (which would require evaluating against human-annotated gold labels, as in Table~\ref{tab:silver_pr}).
The ``x'' symbols denote the operating points used in the end-to-end model, which were chosen based on the number of spans that resulted in the highest ROUGE-2 F$_1$ score on the validation data.

\input{figures/classifier-pr/classifier_pr}

%% file: figures/classifier-pr/classifier_pr.tex
\begin{figure}
    \centering
    \includegraphics[width=\columnwidth]{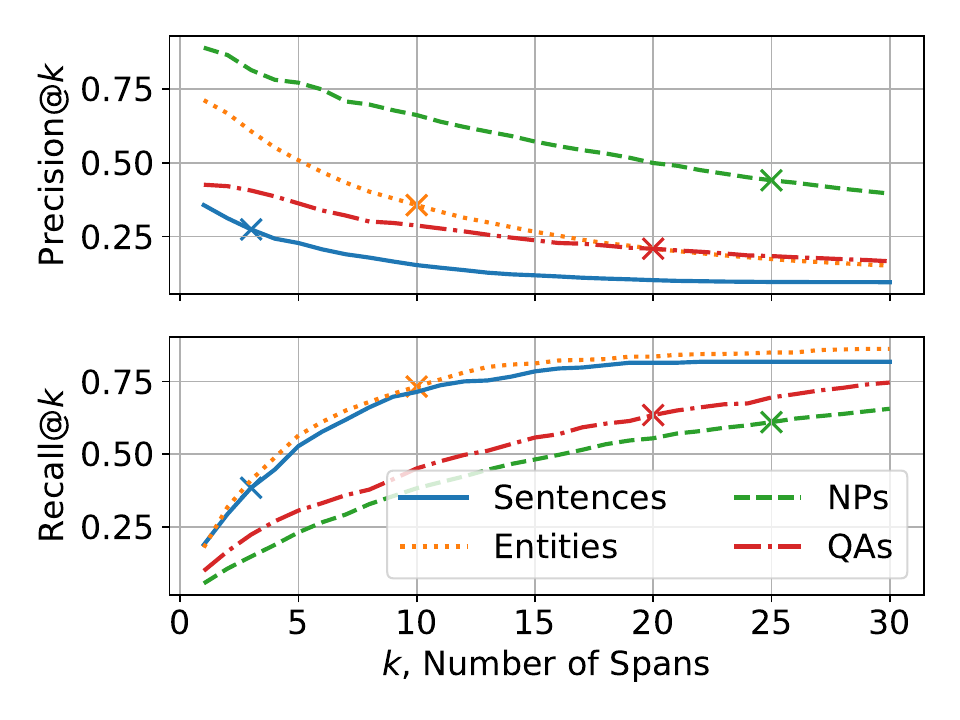}
    \caption{The performances of the salient span classifiers using the different types of salient phrase labeling evaluated against the silver spans.
    The ``x'' marks the operating points used in the end-to-end models.}
    \label{fig:classifier_pr}
\end{figure}

%% file: appendix/xsum.tex
\section{XSum Results}
\label{appendix:xsum}

Table~\ref{tab:xsum_results} contains the automatic summarization evaluation results on the XSum dataset.
These results are included in the Appendix because incorporating the span-based supervision does not improve end-to-end results over the baseline BART model, which is a conclusion also reached by GSum, a model closely related to ours.

\input{figures/xsum_results}

We suspect this is due to the abstractive nature of XSum compared to the more extractive CNN/DailyMail and NYTimes.
Since the methods for identifying salient spans rely on the document and gold summary explicitly stating the salient content, we suspect the abstractiveness of XSum would result in this happening less frequently and thus be less beneficial to a summarization model trained on XSum.

%% file: figures/xsum_results.tex
\begin{table}
    \centering
    \begin{adjustbox}{width=\columnwidth}
    \begin{tabular}{lccccc}
\toprule
\multirow{2}{*}[-0.2em]{\bf Method} & \multicolumn{5}{c}{\bf XSum} \\
\cmidrule{2-6} 
& \bf R1 & \bf R2 & \bf RL & \bf BSc & \bf QAE \\
\midrule
\multicolumn{6}{l}{\emph{Baselines \& Other Work}} \\
BART & 45.1 & 21.3 & 40.9 & - & -\\
BART (ours) & \hphantom{$^\dagger$}45.7$^\dagger$ & \hphantom{$^\dagger$}22.4$^\dagger$ & \hphantom{$^\dagger$}37.2$^\dagger$ & \hphantom{$^\dagger$}91.3$^\dagger$ & \hphantom{$^\dagger$}18.9$^\dagger$\\
GSum & 44.9 & 21.2 & 36.0 & 90.4 & 17.9\\
\midrule
\multicolumn{6}{l}{\emph{Silver Spans}} \\
Sentences & 47.3 & 24.2 & 38.7 & 91.5 & 19.9\\
Entities & 48.1 & 24.2 & 39.1 & 91.7 & 21.3\\
Lexical NPs & \bf 54.3 & \bf 29.3 & \bf 44.1 & \bf 92.4 & \bf 26.1\\
QAs & 47.9 & 24.1 & 39.2 & 91.6 & 21.4\\
\midrule
\multicolumn{6}{l}{\emph{End-to-End}} \\
Sentences & \bf 45.0 & \bf 21.7 & \bf 36.6 & \bf 91.2 & \bf 18.6\\
Entities & 44.1 & 20.9 & 35.9 & 91.0 & 17.6\\
Lexical NPs & 42.5 & 19.2 & 34.2 & 90.8 & 16.4\\
QAs & \bf 45.1 & \bf 21.8 & \bf 36.7 & 91.2 & 17.9\\
\bottomrule
\end{tabular}
\end{adjustbox}
    \caption{The results of the models trained on the XSum dataset as evaluated with the automatic evaluation metrics.
    The span-based models do not improve over the baseline BART, potentially due to the abstractive nature of the XSum dataset.}
    \label{tab:xsum_results}
\end{table}

%% file: appendix/gold_span_annotation.tex
\section{Gold Span Annotation Protocol}
\label{appendix:gold_span}

We selected 50 test instances from the CNN/DailyMail dataset uniformly at random and labeled each of the document NPs as salient or not salient based on whether the corresponding predicate-argument relation also appears in the gold summary.
We did not mark instances in which the NP's predicate-argument relation could be inferred from the gold summary via entailment as salient since our silver span labeling methods aim to mark phrases as salient if the content is explicilty included in the gold summary.

In general, this procedure was straightforward due to the extractive nature of the dataset in which the gold summaries copy heavily from the input document.
If information was repeated in the input document, we tried to label the occurrence which contained the most predicate-argument relations which also matched the gold summary.
That is, we selected the ``best match.''
Otherwise, the first occurrence was selected.

Although our labeling procedure may be noisy, we do not have reason to believe that the labels may be biased in favor of either the lexical NP or QA labeling methods.
Therefore, the statistics calculated from these labels should only be used as diagnostic tools to make relative comparisons between the different labeling methods rather than precise estimates of their exact values.
50 documents were sufficient to achieve statistically different results.

Our annotations will be released after publication.

%% file: appendix/augmented_comparison.tex
\section{Data-Augmentation Automatic Evaluation}
Table~\ref{tab:augmentation_auto_results} contains the comparison between the standard and data-augmented training procedures based on the automatic metrics.
The scores are nearly the same.
The benefit of the model trained on the augmented data is in its controllability, which is not captured by this evaluation because the models trained with the standard and augmented training data receive the same spans as input supervision.

\input{figures/augmentation_automatic_results}

%% file: figures/augmentation_automatic_results.tex
\begin{table}
    \centering
    \begin{adjustbox}{width=\columnwidth}
    \begin{tabular}{lccccc}
        \toprule
        \bf{Model} & \bf{R1} & \bf{R2} & \bf{RL} & \bf{BSc} & \bf{QAE}  \\
        \midrule
        \multicolumn{6}{l}{\emph{Silver Spans}} \\
        QAs & 55.3 & 31.4 & 51.9 & 90.0 & 33.7 \\
        QAs + Data Aug. & 55.2 & 31.3 & 51.7 & 89.9 & 33.4 \\
        \midrule
        \multicolumn{6}{l}{\emph{End-to-End}} \\
        QAs & 45.5 & 21.9 & 42.4 & 88.5 & 24.4 \\
        QAs + Data Aug. & 45.3 & 21.8 & 42.1 & 88.4 & 24.3 \\
        \bottomrule
    \end{tabular}
    \end{adjustbox}
    \caption{The automatic evaluation metrics for summary quality are nearly the same for the QA-based model and the QA-based model trained on the augmented data.}
    \label{tab:augmentation_auto_results}
\end{table}

%% file: appendix/human_eval_details.tex
\section{Human Evaluation Details}
\label{app:human_eval}
Fig.~\ref{fig:hitapp} contains a screenshot of the tool we used for annotating summary quality on MTurk.
The annotators were instructed to rate the summaries from ``Very Poor'' to ``Very Good'' based on whether the summary contained important information, was faithful to the input document, was fluent, and was cohesive.
The ratings were converted to a Likert scale from 1-5 and averaged across all of the ratings for a system.

In order to encourage the annotators to pay attention to the task, we also required that they write a very brief explanation of how they made their decision, inspired by \citet{NZMSNM21}.

The MTurk annotators were paid at a rate of around \$15 USD per hour.

\input{figures/quality-tool-screenshot/quality_tool_screenshot}

%% file: figures/quality-tool-screenshot/quality_tool_screenshot.tex
\begin{figure*}
    \centering
    \includegraphics[width=\textwidth]{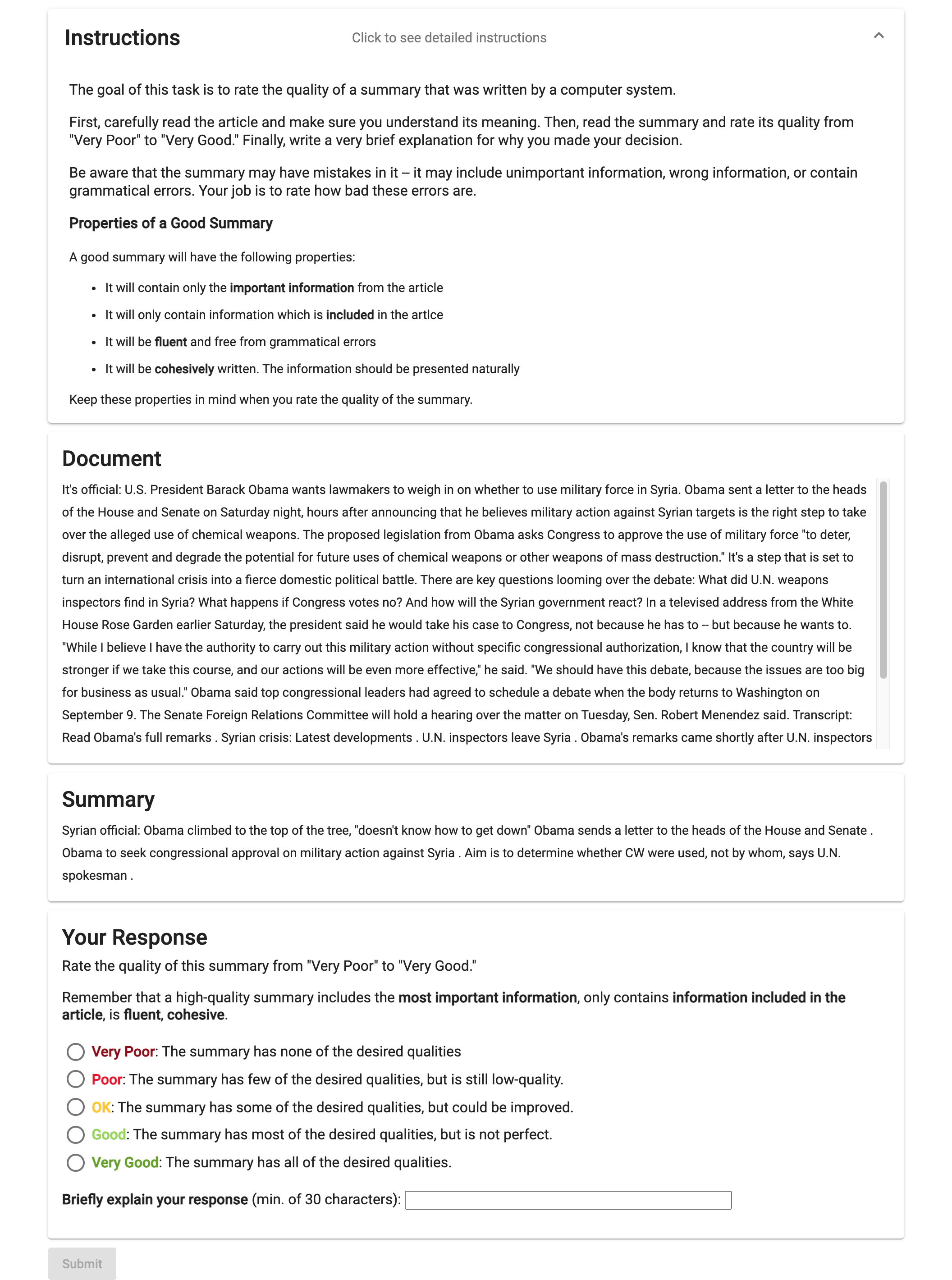}
    \caption{A screenshot of the tool we used for annotating summary quality on MTurk.}
    \label{fig:hitapp}
\end{figure*}

%% file: main.bbl
\begin{thebibliography}{36}
\expandafter\ifx\csname natexlab\endcsname\relax\def\natexlab#1{#1}\fi

\bibitem[{Amplayo et~al.(2021)Amplayo, Angelidis, and Lapata}]{AmplayoAnLa21}
Reinald~Kim Amplayo, Stefanos Angelidis, and Mirella Lapata. 2021.
\newblock \href {https://doi.org/10.18653/v1/2021.emnlp-main.528}
  {{Aspect-Controllable Opinion Summarization}}.
\newblock In \emph{Proceedings of the 2021 Conference on Empirical Methods in
  Natural Language Processing}, pages 6578--6593, Online and Punta Cana,
  Dominican Republic. Association for Computational Linguistics.

\bibitem[{Arumae and Liu(2018)}]{ArumaeLi18}
Kristjan Arumae and Fei Liu. 2018.
\newblock \href {https://doi.org/10.18653/v1/P18-3015} {{Reinforced Extractive
  Summarization with Question-Focused Rewards}}.
\newblock In \emph{Proceedings of {ACL} 2018, Student Research Workshop}, pages
  105--111, Melbourne, Australia. Association for Computational Linguistics.

\bibitem[{Arumae and Liu(2019)}]{ArumaeLi19}
Kristjan Arumae and Fei Liu. 2019.
\newblock \href {https://doi.org/10.18653/v1/N19-1264} {{Guiding Extractive
  Summarization with Question-Answering Rewards}}.
\newblock In \emph{Proceedings of the 2019 Conference of the North {A}merican
  Chapter of the Association for Computational Linguistics: Human Language
  Technologies, Volume 1 (Long and Short Papers)}, pages 2566--2577,
  Minneapolis, Minnesota. Association for Computational Linguistics.

\bibitem[{Cao et~al.(2018)Cao, Wei, Li, and Li}]{CWLL18}
Ziqiang Cao, Furu Wei, Wenjie Li, and Sujian Li. 2018.
\newblock {Faithful to the Original: Fact Aware Neural Abstractive
  Summarization}.
\newblock In \emph{Proceedings of the AAAI Conference on Artificial
  Intelligence}, volume~32.

\bibitem[{Chen and Bansal(2018)}]{ChenBa18}
Yen-Chun Chen and Mohit Bansal. 2018.
\newblock \href {https://doi.org/10.18653/v1/P18-1063} {{Fast Abstractive
  Summarization with Reinforce-Selected Sentence Rewriting}}.
\newblock In \emph{Proceedings of the 56th Annual Meeting of the Association
  for Computational Linguistics (Volume 1: Long Papers)}, pages 675--686,
  Melbourne, Australia. Association for Computational Linguistics.

\bibitem[{Clark et~al.(2020)Clark, Luong, Le, and Manning}]{CLLM20}
Kevin Clark, Minh{-}Thang Luong, Quoc~V. Le, and Christopher~D. Manning. 2020.
\newblock \href {https://openreview.net/forum?id=r1xMH1BtvB} {{{ELECTRA:
  Pre-training Text Encoders as Discriminators Rather Than Generators}}}.
\newblock In \emph{8th International Conference on Learning Representations,
  {ICLR} 2020, Addis Ababa, Ethiopia, April 26-30, 2020}. OpenReview.net.

\bibitem[{Demszky et~al.(2018)Demszky, Guu, and Liang}]{DemszkyGuLi18}
Dorottya Demszky, Kelvin Guu, and Percy Liang. 2018.
\newblock {Transforming Question Answering Datasets Into Natural Language
  Inference Datasets}.
\newblock \emph{ArXiv}, abs/1809.02922.

\bibitem[{Deutsch et~al.(2021)Deutsch, Bedrax-Weiss, and Roth}]{DeutschBeRo21}
Daniel Deutsch, Tania Bedrax-Weiss, and Dan Roth. 2021.
\newblock \href {https://doi.org/10.1162/tacl_a_00397} {{{Towards
  Question-Answering as an Automatic Metric for Evaluating the Content Quality
  of a Summary}}}.
\newblock \emph{Transactions of the Association for Computational Linguistics},
  9:774--789.

\bibitem[{Devlin et~al.(2019)Devlin, Chang, Lee, and Toutanova}]{DCLT19}
Jacob Devlin, Ming-Wei Chang, Kenton Lee, and Kristina Toutanova. 2019.
\newblock \href {https://doi.org/10.18653/v1/N19-1423} {{{BERT}: Pre-training
  of Deep Bidirectional Transformers for Language Understanding}}.
\newblock In \emph{Proceedings of the 2019 Conference of the North {A}merican
  Chapter of the Association for Computational Linguistics: Human Language
  Technologies, Volume 1 (Long and Short Papers)}, pages 4171--4186,
  Minneapolis, Minnesota. Association for Computational Linguistics.

\bibitem[{Dou et~al.(2021)Dou, Liu, Hayashi, Jiang, and Neubig}]{DLHJN21}
Zi-Yi Dou, Pengfei Liu, Hiroaki Hayashi, Zhengbao Jiang, and Graham Neubig.
  2021.
\newblock \href {https://doi.org/10.18653/v1/2021.naacl-main.384} {{{GS}um: A
  General Framework for Guided Neural Abstractive Summarization}}.
\newblock In \emph{Proceedings of the 2021 Conference of the North American
  Chapter of the Association for Computational Linguistics: Human Language
  Technologies}, pages 4830--4842, Online. Association for Computational
  Linguistics.

\bibitem[{Durmus et~al.(2020)Durmus, He, and Diab}]{DurmusHeDi20}
Esin Durmus, He~He, and Mona Diab. 2020.
\newblock \href {https://doi.org/10.18653/v1/2020.acl-main.454} {{{FEQA}: A
  Question Answering Evaluation Framework for Faithfulness Assessment in
  Abstractive Summarization}}.
\newblock In \emph{Proceedings of the 58th Annual Meeting of the Association
  for Computational Linguistics}, pages 5055--5070, Online. Association for
  Computational Linguistics.

\bibitem[{Eyal et~al.(2019)Eyal, Baumel, and Elhadad}]{EyalBaEl19}
Matan Eyal, Tal Baumel, and Michael Elhadad. 2019.
\newblock \href {https://doi.org/10.18653/v1/N19-1395} {{Question Answering as
  an Automatic Evaluation Metric for News Article Summarization}}.
\newblock In \emph{Proceedings of the 2019 Conference of the North {A}merican
  Chapter of the Association for Computational Linguistics: Human Language
  Technologies, Volume 1 (Long and Short Papers)}, pages 3938--3948,
  Minneapolis, Minnesota. Association for Computational Linguistics.

\bibitem[{Fan et~al.(2018)Fan, Grangier, and Auli}]{FanGrAu18}
Angela Fan, David Grangier, and Michael Auli. 2018.
\newblock \href {https://doi.org/10.18653/v1/W18-2706} {{Controllable
  Abstractive Summarization}}.
\newblock In \emph{Proceedings of the 2nd Workshop on Neural Machine
  Translation and Generation}, pages 45--54, Melbourne, Australia. Association
  for Computational Linguistics.

\bibitem[{He et~al.(2020)He, Kryściński, McCann, Rajani, and Xiong}]{HKMRX20}
Junxian He, Wojciech Kryściński, Bryan McCann, Nazneen Rajani, and Caiming
  Xiong. 2020.
\newblock \href {http://arxiv.org/abs/2012.04281} {{CTRLsum: Towards Generic
  Controllable Text Summarization}}.

\bibitem[{Jin et~al.(2020)Jin, Wang, and Wan}]{JinWaWa20}
Hanqi Jin, Tianming Wang, and Xiaojun Wan. 2020.
\newblock {SemSUM: Semantic Dependency Guided Neural Abstractive
  Summarization}.
\newblock In \emph{Proc. of the Conference on Artificial Intelligence (AAAI)}.

\bibitem[{Lewis et~al.(2020)Lewis, Liu, Goyal, Ghazvininejad, Mohamed, Levy,
  Stoyanov, and Zettlemoyer}]{LLGGMLSZ20}
Mike Lewis, Yinhan Liu, Naman Goyal, Marjan Ghazvininejad, Abdelrahman Mohamed,
  Omer Levy, Veselin Stoyanov, and Luke Zettlemoyer. 2020.
\newblock \href {https://doi.org/10.18653/v1/2020.acl-main.703} {{{BART}:
  Denoising Sequence-to-Sequence Pre-training for Natural Language Generation,
  Translation, and Comprehension}}.
\newblock In \emph{Proceedings of the 58th Annual Meeting of the Association
  for Computational Linguistics}, pages 7871--7880, Online. Association for
  Computational Linguistics.

\bibitem[{Lin(2004)}]{Lin04}
Chin-Yew Lin. 2004.
\newblock \href {https://www.aclweb.org/anthology/W04-1013} {{{ROUGE: A Package
  for Automatic Evaluation of Summaries}}}.
\newblock In \emph{Text Summarization Branches Out}, pages 74--81, Barcelona,
  Spain. Association for Computational Linguistics.

\bibitem[{Loshchilov and Hutter(2019)}]{LoshchilovHu19}
Ilya Loshchilov and Frank Hutter. 2019.
\newblock {Decoupled Weight Decay Regularization}.
\newblock In \emph{Proc. of the International Conference on Learning
  Representations}.

\bibitem[{Matsumaru et~al.(2020)Matsumaru, Takase, and
  Okazaki}]{MatsumaruTaOk20}
Kazuki Matsumaru, Sho Takase, and Naoaki Okazaki. 2020.
\newblock \href {https://doi.org/10.18653/v1/2020.acl-main.123} {{Improving
  Truthfulness of Headline Generation}}.
\newblock In \emph{Proceedings of the 58th Annual Meeting of the Association
  for Computational Linguistics}, pages 1335--1346, Online. Association for
  Computational Linguistics.

\bibitem[{Nallapati et~al.(2017)Nallapati, Zhai, and Zhou}]{NallapatiZhZh17}
Ramesh Nallapati, Feifei Zhai, and Bowen Zhou. 2017.
\newblock {SummaRuNNer: A Recurrent Neural Network Based Sequence Model for
  Extractive Summarization of Documents}.
\newblock In \emph{Proc. of the Conference on Artificial Intelligence (AAAI)}.

\bibitem[{Nallapati et~al.(2016)Nallapati, Zhou, dos Santos, Gul{\c{c}}ehre,
  and Xiang}]{NZSGX16}
Ramesh Nallapati, Bowen Zhou, Cicero dos Santos, {\c{C}}a{\u{g}}lar
  Gul{\c{c}}ehre, and Bing Xiang. 2016.
\newblock \href {https://doi.org/10.18653/v1/K16-1028} {{Abstractive Text
  Summarization using Sequence-to-sequence {RNN}s and Beyond}}.
\newblock In \emph{Proceedings of The 20th {SIGNLL} Conference on Computational
  Natural Language Learning}, pages 280--290, Berlin, Germany. Association for
  Computational Linguistics.

\bibitem[{Nan et~al.(2021)Nan, Nallapati, Wang, dos Santos, Zhu, Zhang,
  McKeown, and Xiang}]{NNWSZZMX21}
Feng Nan, Ramesh Nallapati, Zhiguo Wang, Cicero~Nogueira dos Santos, Henghui
  Zhu, Dejiao Zhang, Kathleen McKeown, and Bing Xiang. 2021.
\newblock \href {https://aclanthology.org/2021.eacl-main.235} {{Entity-level
  Factual Consistency of Abstractive Text Summarization}}.
\newblock In \emph{Proceedings of the 16th Conference of the European Chapter
  of the Association for Computational Linguistics: Main Volume}, pages
  2727--2733, Online. Association for Computational Linguistics.

\bibitem[{Narayan et~al.(2018)Narayan, Cohen, and Lapata}]{NarayanCoLa18a}
Shashi Narayan, Shay~B. Cohen, and Mirella Lapata. 2018.
\newblock \href {https://doi.org/10.18653/v1/D18-1206} {{Don't Give Me the
  Details, Just the Summary! Topic-Aware Convolutional Neural Networks for
  Extreme Summarization}}.
\newblock In \emph{Proceedings of the 2018 Conference on Empirical Methods in
  Natural Language Processing}, pages 1797--1807, Brussels, Belgium.
  Association for Computational Linguistics.

\bibitem[{Narayan et~al.(2021)Narayan, Zhao, Maynez, Simoes, Nikolaev, and
  McDonald}]{NZMSNM21}
Shashi Narayan, Yao-Dong Zhao, Joshua Maynez, Gonccalo Simoes, Vitaly Nikolaev,
  and Ryan~T. McDonald. 2021.
\newblock \href {https://arxiv.org/abs/2104.07606} {{Planning with Learned
  Entity Prompts for Abstractive Summarization}}.

\bibitem[{Puduppully et~al.(2019)Puduppully, Dong, and
  Lapata}]{PuduppullyDoLa19}
Ratish Puduppully, Li~Dong, and Mirella Lapata. 2019.
\newblock {Data-to-Text Generation with Content Selection and Planning}.
\newblock In \emph{Proceedings of the AAAI Conference on Artificial
  Intelligence}, volume~33, pages 6908--6915.

\bibitem[{Rajpurkar et~al.(2018)Rajpurkar, Jia, and Liang}]{RajpurkarJiLi18}
Pranav Rajpurkar, Robin Jia, and Percy Liang. 2018.
\newblock \href {https://doi.org/10.18653/v1/P18-2124} {{Know What You Don{'}t
  Know: Unanswerable Questions for {SQ}u{AD}}}.
\newblock In \emph{Proceedings of the 56th Annual Meeting of the Association
  for Computational Linguistics (Volume 2: Short Papers)}, pages 784--789,
  Melbourne, Australia. Association for Computational Linguistics.

\bibitem[{Rush et~al.(2015)Rush, Chopra, and Weston}]{RMCW15}
Alexander~M. Rush, Sumit Chopra, and Jason Weston. 2015.
\newblock \href {https://doi.org/10.18653/v1/D15-1044} {{A Neural Attention
  Model for Abstractive Sentence Summarization}}.
\newblock In \emph{Proceedings of the 2015 Conference on Empirical Methods in
  Natural Language Processing}, pages 379--389, Lisbon, Portugal. Association
  for Computational Linguistics.

\bibitem[{Sandhaus(2008)}]{Sandhaus08}
Evan Sandhaus. 2008.
\newblock {The New York Times Annotated Corpus}.
\newblock \emph{Linguistic Data Consortium, Philadelphia}, 6(12):e26752.

\bibitem[{Scialom et~al.(2021)Scialom, Dray, Lamprier, Piwowarski, Staiano,
  Wang, and Gallinari}]{SDLPSWG21}
Thomas Scialom, Paul-Alexis Dray, Sylvain Lamprier, Benjamin Piwowarski, Jacopo
  Staiano, Alex Wang, and Patrick Gallinari. 2021.
\newblock \href {https://aclanthology.org/2021.emnlp-main.529} {{{Q}uest{E}val:
  Summarization Asks for Fact-based Evaluation}}.
\newblock In \emph{Proceedings of the 2021 Conference on Empirical Methods in
  Natural Language Processing}, pages 6594--6604, Online and Punta Cana,
  Dominican Republic. Association for Computational Linguistics.

\bibitem[{Scialom et~al.(2019)Scialom, Lamprier, Piwowarski, and
  Staiano}]{SLPS19}
Thomas Scialom, Sylvain Lamprier, Benjamin Piwowarski, and Jacopo Staiano.
  2019.
\newblock \href {https://doi.org/10.18653/v1/D19-1320} {{Answers Unite!
  Unsupervised Metrics for Reinforced Summarization Models}}.
\newblock In \emph{Proceedings of the 2019 Conference on Empirical Methods in
  Natural Language Processing and the 9th International Joint Conference on
  Natural Language Processing (EMNLP-IJCNLP)}, pages 3246--3256, Hong Kong,
  China. Association for Computational Linguistics.

\bibitem[{Shapira et~al.(2017)Shapira, Ronen, Adler, Amsterdamer, Bar-Ilan, and
  Dagan}]{SRAABD17}
Ori Shapira, Hadar Ronen, Meni Adler, Yael Amsterdamer, Judit Bar-Ilan, and Ido
  Dagan. 2017.
\newblock \href {https://doi.org/10.18653/v1/D17-2019} {{Interactive
  Abstractive Summarization for Event News Tweets}}.
\newblock In \emph{Proceedings of the 2017 Conference on Empirical Methods in
  Natural Language Processing: System Demonstrations}, pages 109--114,
  Copenhagen, Denmark. Association for Computational Linguistics.

\bibitem[{Wang et~al.(2020)Wang, Cho, and Lewis}]{WangChLe20}
Alex Wang, Kyunghyun Cho, and Mike Lewis. 2020.
\newblock \href {https://doi.org/10.18653/v1/2020.acl-main.450} {{Asking and
  Answering Questions to Evaluate the Factual Consistency of Summaries}}.
\newblock In \emph{Proceedings of the 58th Annual Meeting of the Association
  for Computational Linguistics}, pages 5008--5020, Online. Association for
  Computational Linguistics.

\bibitem[{Wei and Jia(2021)}]{WeiJi21}
Johnny Wei and Robin Jia. 2021.
\newblock \href {https://doi.org/10.18653/v1/2021.acl-long.533} {{The
  Statistical Advantage of Automatic NLG Metrics at the System Level}}.
\newblock In \emph{Proceedings of the 59th Annual Meeting of the Association
  for Computational Linguistics and the 11th International Joint Conference on
  Natural Language Processing (Volume 1: Long Papers)}, pages 6840--6854,
  Online. Association for Computational Linguistics.

\bibitem[{Weiss et~al.(2021)Weiss, Roit, Klein, Ernst, and Dagan}]{WRKED21}
Daniela~Brook Weiss, Paul Roit, Ayal Klein, Ori Ernst, and Ido Dagan. 2021.
\newblock \href {https://aclanthology.org/2021.emnlp-main.778} {{{QA}-Align:
  Representing Cross-Text Content Overlap by Aligning Question-Answer
  Propositions}}.
\newblock In \emph{Proceedings of the 2021 Conference on Empirical Methods in
  Natural Language Processing}, pages 9879--9894, Online and Punta Cana,
  Dominican Republic. Association for Computational Linguistics.

\bibitem[{Zhang et~al.(2020)Zhang, Kishore, Wu, Weinberger, and
  Artzi}]{ZKWWA20}
Tianyi Zhang, Varsha Kishore, Felix Wu, Kilian~Q. Weinberger, and Yoav Artzi.
  2020.
\newblock \href {https://openreview.net/forum?id=SkeHuCVFDr} {{BERTScore:
  Evaluating Text Generation with BERT}}.
\newblock In \emph{8th International Conference on Learning Representations,
  {ICLR} 2020, Addis Ababa, Ethiopia, April 26-30, 2020}. OpenReview.net.

\bibitem[{Zhu et~al.(2021)Zhu, Hinthorn, Xu, Zeng, Zeng, Huang, and
  Jiang}]{ZHXZZHJ21}
Chenguang Zhu, William Hinthorn, Ruochen Xu, Qingkai Zeng, Michael Zeng,
  Xuedong Huang, and Meng Jiang. 2021.
\newblock \href {https://doi.org/10.18653/v1/2021.naacl-main.58} {{Enhancing
  Factual Consistency of Abstractive Summarization}}.
\newblock In \emph{Proceedings of the 2021 Conference of the North American
  Chapter of the Association for Computational Linguistics: Human Language
  Technologies}, pages 718--733, Online. Association for Computational
  Linguistics.

\end{thebibliography}
